# An interpretable data-driven approach to optimizing clinical fall risk assessment

Fardin Ganjkhanloo[1,2*], Emmett Springer[2,3*], Erik H. Hoyer[4,5], Daniel L. Young[4,6], Holley Farley[7], Kimia Ghobadi[2,3]

∗F. Ganjkhanloo and E. Springer contributed equally to this work. This work was supported by the Doctors Company Foundation. Corresponding author: K. Ghobadi.
[1]Center for Health Systems and Policy Modeling, Department of Health Policy and Management, Johns Hopkins University, Baltimore, MD, USA. Email: fganjkh1@jhu.edu
[2]Malone Center for Engineering in Healthcare, Johns Hopkins University, Baltimore, MD, USA
[3]Center for Systems Science and Engineering, Department of Civil and Systems Engineering, Johns Hopkins University, Baltimore, MD, USA. Emails: espring6@jh.edu, kimia@jhu.edu
[4]Department of Physical Medicine and Rehabilitation, School of Medicine, Johns Hopkins University, Baltimore, MD, USA.
[5]Johns Hopkins Hospital, Baltimore, MD, USA. Email: ehoyer1@jhmi.edu
[6]Department of Physical Therapy, University of Nevada, Las Vegas, Las Vegas, NV, USA. Email: daniel.young@unlv.edu
7 Department of Nursing, The Johns Hopkins Hospital. Email: hmcrann1@jhmi.edu

## Abstract

In this study, we aim to better align fall risk prediction from the Johns Hopkins Fall Risk Assessment Tool (JHFRAT) with additional clinically meaningful measures via a data-driven modelling approach. We conducted a retrospective cohort analysis of 54,209 inpatient admissions from three Johns Hopkins Health System hospitals between March 2022 and October 2023. A total of 20,208 admissions were included as high fall risk encounters, and 13,941 were included as low fall risk encounters. To incorporate clinical knowledge and maintain interpretability, we employed constrained score optimization (CSO) models to reweight the JHFRAT scoring weights, while preserving its additive structure and clinical thresholds. Recalibration refers to adjusting item weights so that the resulting score can order encounters more consistently by the study's risk labels, and without changing the tool's form factor or deployment workflow. The model demonstrated significant improvements in predictive performance over the current JHFRAT (CSO AUC-ROC=0.91, JHFRAT AUC-ROC=0.86). This performance improvement translates to protecting an additional 35 high-risk patients per week across the Johns Hopkins Health System. The constrained score optimization models performed similarly with and without the EHR variables. Although the benchmark black-box model (XGBoost), improves upon the performance metrics of the knowledge-based constrained logistic regression (AUC-ROC=0.94), the CSO demonstrates more robustness to variations in risk labelling. This evidence-based approach provides a robust foundation for health systems to systematically enhance inpatient fall prevention protocols and patient safety using data-driven optimization techniques, contributing to improved risk assessment and resource allocation in healthcare settings.

## 1. Introduction

Inpatient falls remain a critical issue within hospital settings, leading to increased morbidity, mortality, and significant healthcare costs.[1] Fall-related injuries, particularly among older adults, result in prolonged hospital stays, reduced quality of life, and substantial resource demands on healthcare systems.[2,3] More broadly, falls represent a prototypical hospital-acquired harm for which prevention depends on accurate risk stratification, timely clinical decision-making, and judicious allocation of limited resources. Widespread implementation of intensive risk-mitigation interventions, such as increased rounding and restriction of patient mobility, is resource-intensive and may offer limited value for patients with low fall risk.[4,5] At the same time, underestimating or failing to identify patients truly at high risk can result in missed opportunities for timely intervention and prevention of serious harm. Accurate fall risk assessment is essential for implementing targeted prevention strategies and optimizing resource allocation, a challenge shared across many hospital safety domains where adverse events are uncommon but consequences are substantial.

However, developing reliable models of fall risk is challenging, as falls are relatively rare events and true underlying risk is difficult to observe directly. Because fall prevention interventions are often applied preemptively to those perceived as high risk, the resulting data often reflect patterns of intervention rather than the unmitigated likelihood of falling. This phenomenon, where observed outcomes are shaped by preventive actions, complicates risk modeling for many hospital-acquired conditions and obscures direct measurement of patient-level vulnerability. Yet without methods that can disentangle true fall risk from the effects of preventive interventions, our understanding of patient-level vulnerability remains incomplete, a critical gap that must be addressed to improve both research validity and clinical decision-making.

The Johns Hopkins Fall Risk Assessment Tool (JHFRAT)[6] is widely used in hospitals to score fall risk based on seven domains assessed by nursing staff once per shift and as patient condition changes. However, its scoring system and risk thresholds were developed using an evidence-based approach[7], not empirically derived, raising concerns about their ability to represent the complex, dynamic nature of fall risk. Furthermore, the reliance on clinical judgement and absence of an empirical ground truth for an individual's true fall risk complicates validation and model refinement. While some previous research utilizes "off-the-shelf" machine learning models to predict inpatient fall risk[8-12], few address the risk-obscuring effects of interventions[13] or seek to optimize existing clinical assessments with a focus on interpretability and assuredness, features that are critical for decision support tools intended for frontline use across hospital safety domains.

In this study, we use a "score-based recalibration" by adjusting the relative weights assigned to JHFRAT items (and their implied category separation) so that higher total scores correspond to higher risk labels more consistently. Our goal is to improve the ranking and category assignment of an existing clinical score, while preserving its interpretability and clinical use pattern by (1) developing a clinically informed proxy for true fall risk based on observed fall prevention practices; (2) applying a constrained score optimization framework to recalibrate JHFRAT coefficients and thresholds while maintaining interpretability; and (3) evaluating whether incorporating additional EHR-derived functional and clinical variables further improves fall risk prediction by developing augmented JHFRAT models. Through this multi-site analysis, we aimed to determine whether data-driven modeling could enhance the accuracy and clinical utility of an established risk assessment tool without compromising its transparency or ease of implementation. The methods presented have broad applicability to risk assessment in other healthcare contexts where adverse events are rare or ground-truth risk is greatly obscured by intervention. Although falls serve as the motivating use case, the methods presented provide a proof of concept for improving risk assessment in other hospital-acquired harms characterized by rare outcomes, preventive interventions, and decision-dependent data generation.

## 2. Methods

### 2.1 Data Source

We follow the TRIPOD and STROBE statement guidelines for reporting the study data source and methodology.[14] Data for this study was extracted from electronic health records (EHR) for all adult non-psychiatric inpatient encounters with admission and discharge between March 28th 2022 and November 3rd 2023 across three Johns Hopkins Health System hospitals: Johns Hopkins Hospital (JHH), Johns Hopkins Bayview Medical Center (BMC), and Howard County Medical Center (HCM). A total of 54,207 encounters from the selected period had lengths of stay of at least 48 hours and constitute the study data set. The Johns Hopkins Fall Risk Assessment Tool (JHFRAT) consists of 18 binary variables across seven categories: age, cognition, elimination, fall history, patient care equipment, medication, and mobility. In addition to these 18 variables, nurses are required to record what fall prevention interventions, if any, the patient will receive (or is currently receiving). There are 30 standardized intervention options that can be recorded as binary variables, and we retain the records for these variables in the data source along with each recorded JHFRAT.

The extracted EHR data also includes daily measures of patient mobility with the Johns Hopkins Highest Level of Mobility (JH-HLM)[15] scores and Activity Measure for Post-Acute Care Basic Mobility '6-clicks' Short Form (AM-PAC)[16] scores, ICD-10 diagnostic codes, age, gender, race, service category, and daily intervention records. These additional variables were selected for inclusion in the data source based on established clinical relevance[17-20] to fall risk assessment and data availability within the EHR system. The JHFRAT, AMPAC, and JH-HLM are mandated to be completed once during each 12-hour nursing shift, and the study data includes all completed assessments for each encounter. Demographic variables, the encounter department service category, and ICD-10 codes are documented once per encounter in the data source.

This health system follows the National Database for Nursing Quality Indicators (NDNQI) guidelines for fall and fall with injury data collection and reporting. During the study period, 498 encounters (0.92% of encounters) included at least one fall event documented and later reviewed in accordance with the NDNQI guidelines. There are 623 total documented falls among these encounters, resulting in 1.5 falls per 1,000 patient-days across all encounters in the data. We received approval for the study protocol from the Johns Hopkins Institutional Review Board, and data collection followed established privacy protection guidelines.

### 2.2. Study Design

Any encounters with fewer than three completed JHFRAT records in the data are excluded. We included only encounters with a length of stay between 2 and 21 full days. Furthermore, among the encounters with one or more fall events, we excluded any patients whose first fall was before the 3rd day or after the 21st day of the encounter and excluded all EHR data recorded after the fall.

We conducted structured clinical review sessions with clinical experts from nurse quality and physical medicine and rehabilitation to characterize interventions among the 30 options as standard or targeted, based on clinical expertise. A targeted intervention is defined for the purpose of this study as resource-intensive measures not suitable for broad implementation, but rather reserved for only high fall risk patients. The following 12 interventions were identified as targeted: increased rounding, injury prevention hospital protocol (e.g. bleeding risk, fracture risk), PT/OT consult, frequent reorientation, bed/chair exit alarm, bladder management device, elimination schedule, fall prevention education to caregivers, early delirium recognition and treatment, physical bed restraints, constant observation, and protective devices (e.g. hip protector, floor mat, helmet, low bed). Interventions considered standard include ensuring patient is safe and independent with use of assistive device, keeping equipment on one side of patient, and educating patient to call/wait prior to mobilizing.

We categorized patients who consistently received several or consistently received very few targeted interventions into the following risk groups:

- Low risk: The patient received no more than one targeted intervention per (overlapping) three-day window in a stretch spanning at least half of the encounter.
- High risk: The patient received at least six targeted interventions per (overlapping) three-day window in a stretch spanning at least half of the encounter.
- Indeterminant: Neither low-risk nor high-risk criteria are met

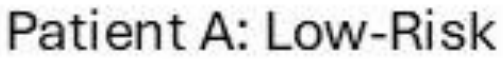

Patient B: High-Risk

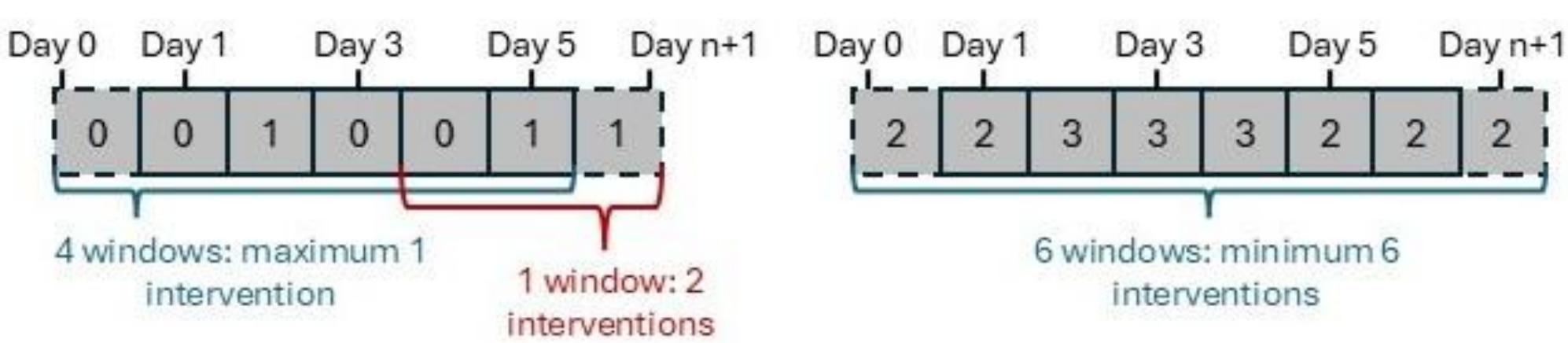

Patient C: Indeterminant

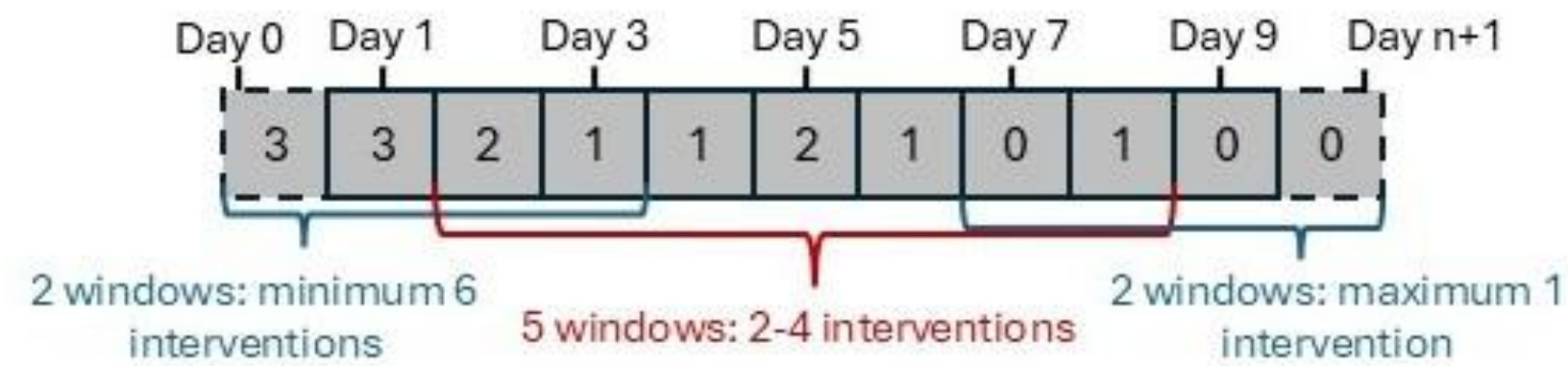

Figure 1: Examples of encounters labelled as low-risk (A), high-risk (B) and indeterminant (C) by targeted intervention-based criteria during overlapping three-day windows.

This approach allowed us to utilize binary classification methods and is a more conservative labelling approach in the face of true fall risk uncertainty. When analyzing the three-day windows, we adjusted for the beginning and end of the encounter by considering the first and last day twice. This is illustrated in Figure 1 by the addition of Day 0, which has the same number of interventions as Day 1, and Day n+1, which has the same number of interventions as Day n. With the addition of these days, each actual day is included in exactly three three-day windows, avoiding biasing the encounter labels towards the days in the middle of the encounter.

We conducted a matching procedure to identify encounters in the indeterminant group with prevention patterns like those preceding falls. An indeterminate encounter was considered high risk via matching if any of its three-day windows demonstrated the same configuration of targeted interventions as the three-day window immediately preceding a fall in a high-risk encounter. For each high-risk fall encounter, up to three matched indeterminate encounters were selected based on the greatest similarity in non-targeted interventions. The full labeling methodology and resulting cohort are illustrated in Figure 2. To evaluate the robustness of each of the models to uncertainty in the data labels, we performed sensitivity varying the minimum number of interventions needed per 3-day window for a patient to qualify as high-risk between four and eight.

### 2.3. Model Implementation

We employ two complementary modeling approaches to both quantify the predictive gains achievable through additional electronic health record (EHR) data and compare the performance of the white-box and black-box models for fall prediction. First, we employed a constrained optimization model that preserved the structure of the current JHFRAT assessment tool while incorporating additional EHR variables. This model is a JHFRAT-specific implementation of the recently developed generalized constrained score optimization (CSO) model for ordinal classification.[21] Let $X \in \mathbb{R}^{n \times m}$ represent the feature matrix for $n$ patients and $m$ risk factors, and $y \in \{0,1\}^n$ denote the binary risk label (low risk = 0, high risk = 1). Encounter-specific objective weights are defined by $w_i$. Let $C$ represent the set of ordered pairs $(j, k)$ encoding clinical hierarchies in scores. The optimization problem seeks to determine coefficient $\beta \in \mathbb{R}^{\mathrm{m}}$ that maximizes predictive accuracy while satisfying clinical constraints via the following formulation:

$$
\begin{aligned}
\mathbf{CSO}(X, y, \lambda)\colon \max_{\beta, s} \quad & \lambda L_{T_1} + (1-\lambda) L_{T_2} \\
\text{s.t.} \quad & L_T = \textstyle\sum_{i=1}^{n} w_i \left( y_i (X_i \beta - T) - \ln\left(1 + e^{X_i \beta - T}\right) \right), \\
& \beta_j \leq \beta_k \quad \forall (j,k) \in C, \\
& \beta_j \geq 0 \quad \forall j \in [m],
\end{aligned}
$$

$$
\text{where } \; w_i = \begin{cases} \dfrac{1}{\sum_{i=1}^{n} y_i} & y_i = 1 \\[2ex] \dfrac{1}{1-\sum_{i=1}^{n} y_i} & y_i = 0 \end{cases}
$$

The risk score for each encounter is defined by $X_i \beta$, maintaining the additive structure of JHFRAT. The optimization objective is a weighted combination of the score log-likelihoods, $L_T$, with two different thresholds: $T_1 = 6$ and $T_2 = 13$ to match the current

JHFRAT category thresholds. Because the objective is optimized at the established JHFRAT cut points, the resulting coefficients can be interpreted as a score recalibration that improves concordance with the study labels while maintaining the original score structure and operational thresholds[6]. The model essentially performed multi-task learning by simultaneously optimizing for two different scenarios: encounters labeled low-risk should encompass the low and moderate-risk categories, while high-risk encounters in the study cohort should only be classified as high-risk, and vice versa. We ran the optimization first with $\lambda = 0.5$, then varied this multi-objective weighting parameter in the sensitivity analysis. The weighted sample normalization via $w_i$ helped address the class imbalance in our dataset by making the total weight per class equal in the objective.

The coefficient ordering constraints incorporated structured clinical knowledge by preserving the ordinal relationships within each of the following single-select JHFRAT categories indicated in Table 1: Age, Medications, and Patient Care Equipment. We therefore ensured that higher levels of assessed risk factors contributed to progressively greater risk scores than lower levels within the same domain. Non-negativity constraints guarantee positive risk contributions. Notably, this optimization architecture readily accommodates additional constraint formulations, enabling systematic incorporation of evolving clinical knowledge or institution-specific requirements.

We train two versions of the CSO models: Optimized CSO, which utilizes only the 18 current JHFRAT risk factors as features, and Augmented CSO, which includes 22 additional binary EHR variables (Table 1), that include demographics, AMPAC and JH-HLM score ranges, and department service category. We additionally trained common models with the best performance achieved by gradient boosted decision tree models (XGBoost). The model was trained on both the JHFRAT-only and expanded datasets to establish empirical performance benchmarks achievable through unconstrained utilization of the expanded variable set. Since each of the JHFRAT assessment components and the JH-HLM and AM-PAC scores are documented multiple times throughout each encounter, we utilized their average value across the encounter for model training and evaluation and do not impute any values for assessments missed in a shift. We split the encounter data into 80% training and 20% testing sets, stratified by risk label. We performed a stratified 5-fold cross-validation to ensure consistent class ratios across folds. All CSO models were solved with the CVXPY optimization library MOSEK solver with a convergence threshold of $10^{-8}$ and $10^{6}$ maximum iterations. All XGBoost models were trained using the Python XGBoost library with hyperparameters (100 estimators, learning rate of 0.1, no maximum depth) selected through preliminary exploration.

## 3. Results

### 3.1. Study Cohort Composition

Applying the intervention-based criteria, 20,265 encounters were categorized as low risk, 13,836 as high risk, and 12,935 as indeterminate. Within the high-risk group, 219 encounters included at least one documented fall. To refine the indeterminate group, we applied the matching procedure described above, identifying 108 indeterminate encounters that shared identical three-day intervention patterns with 45 of the high-risk fall encounters. These matched encounters were reclassified as high risk. The final analytic cohort therefore comprised 20,265 (59.2%) low-risk and 13,945 (40.8%) high-risk encounters, as illustrated in Figure 2. Of the 498 encounters including at least one fall event, 61 were in the excluded group, 108 were in the low-risk group, 221 were in the high-risk group, and the remaining 108 were in the indeterminate group.

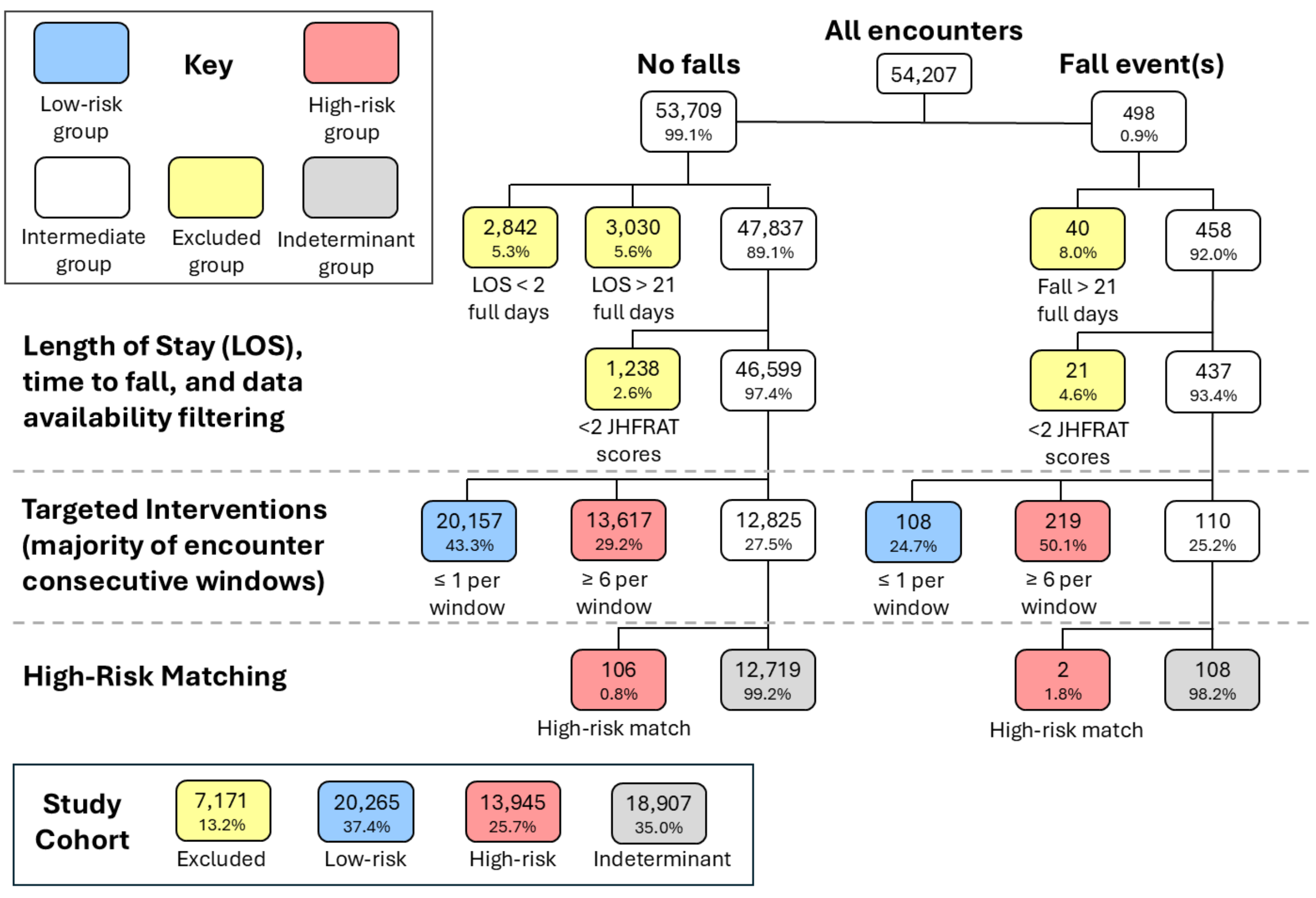

Figure 2. Encounter exclusion and cohort selection steps and outcomes.

The Spearman correlation between average daily JHFRAT score and total daily targeted interventions for all encounters was 0.61 ($p < 0.001$), indicating that patients with higher JHFRAT scores received more of these targeted interventions, as intended. However, Figure 3 demonstrates how the number of interventions varies widely between patients with similar average daily JHFRAT scores despite the clear correlation. Across all encounters, the scores are largely clustered below 10, with fewer than 2 targeted interventions per day on average. When we separate the encounters into the low-risk and high-risk groups in the study cohort, there is a notable difference in the data distributions. The low-risk group is even more highly concentrated towards few daily targeted interventions, while the high-risk group's data clusters between 2 and 6 daily targeted interventions and JHFRAT scores between 5 and 15.

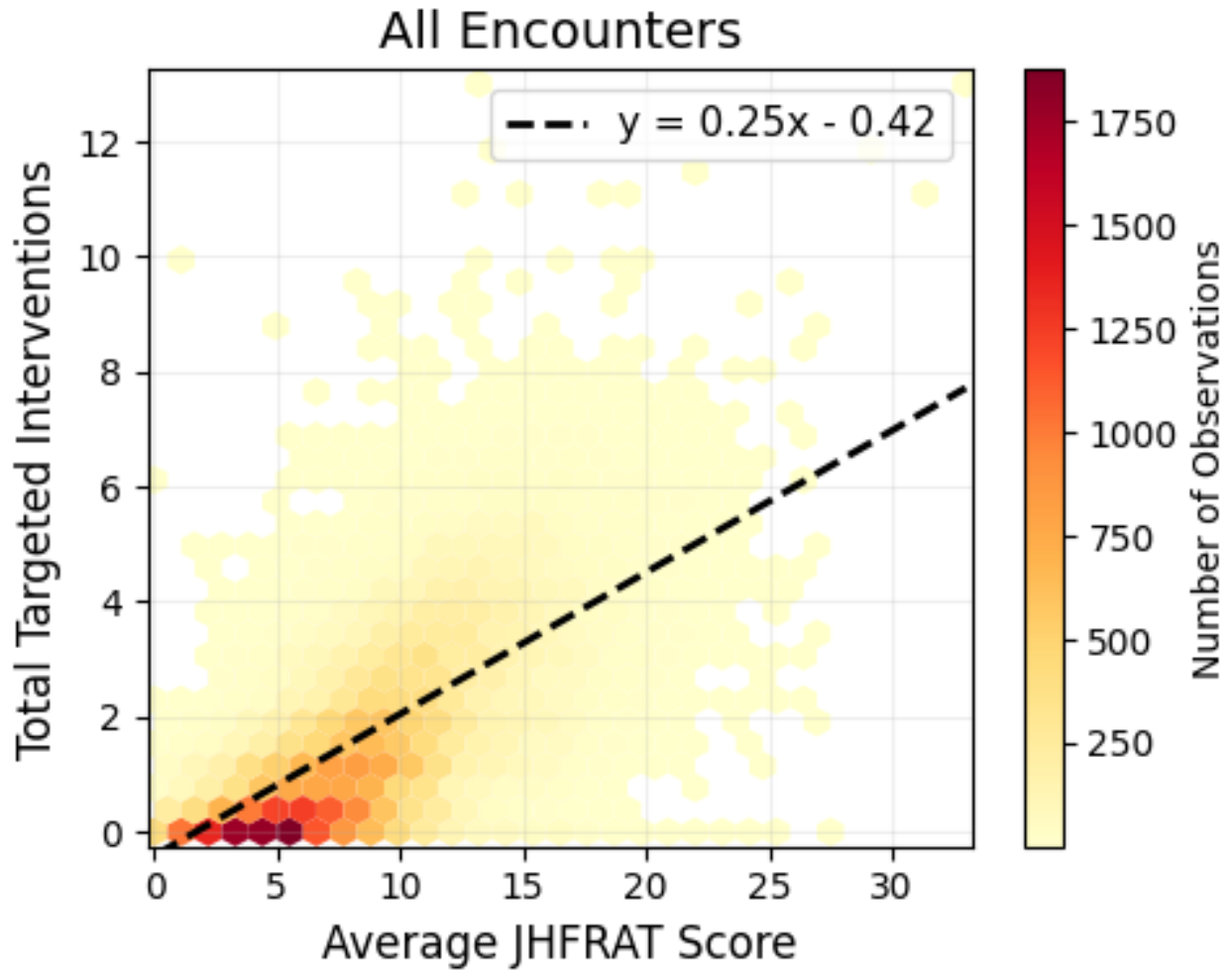

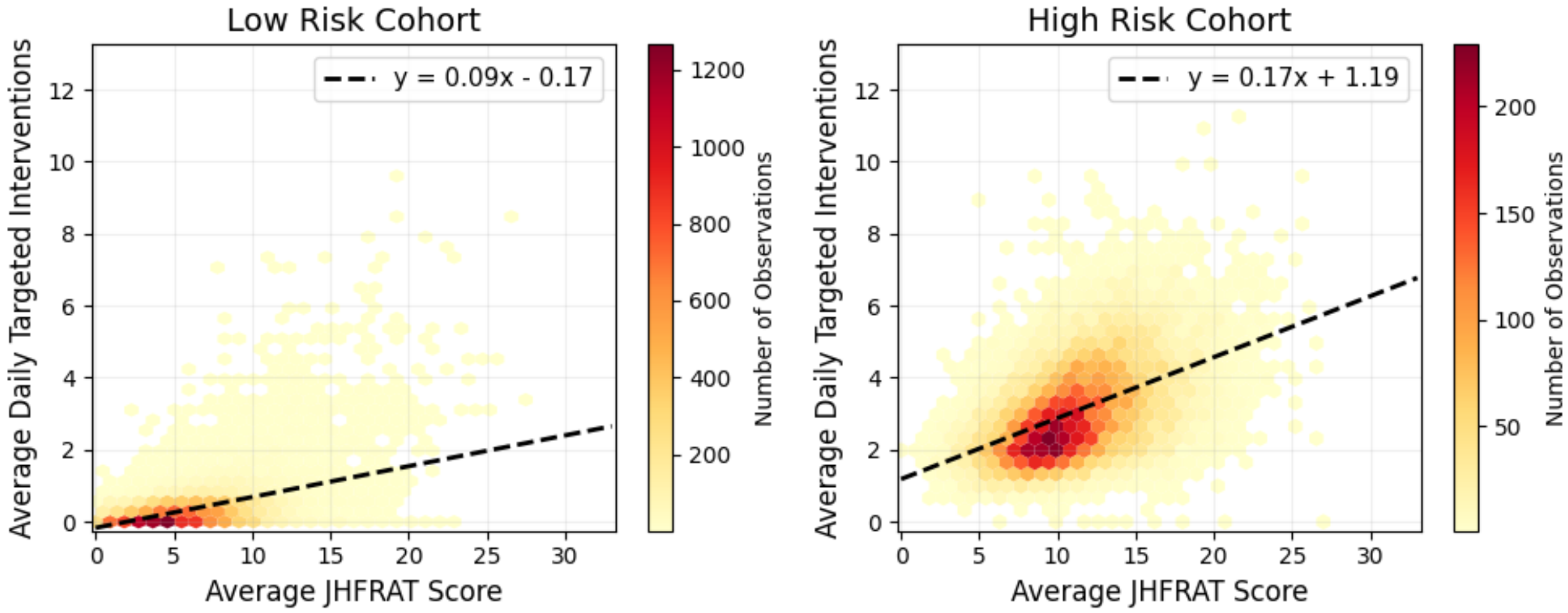

Figure 3. Correlation between encounter average JHFRAT scores and average daily number of targeted interventions for all encounters (above), encounters in the low-risk study cohort (below left) and encounters in the high-risk study cohort (below right). JHFRAT score and average daily targeted interventions are correlated, with both generally higher in the high-risk cohort.

### 3.2. Model Performance Comparison

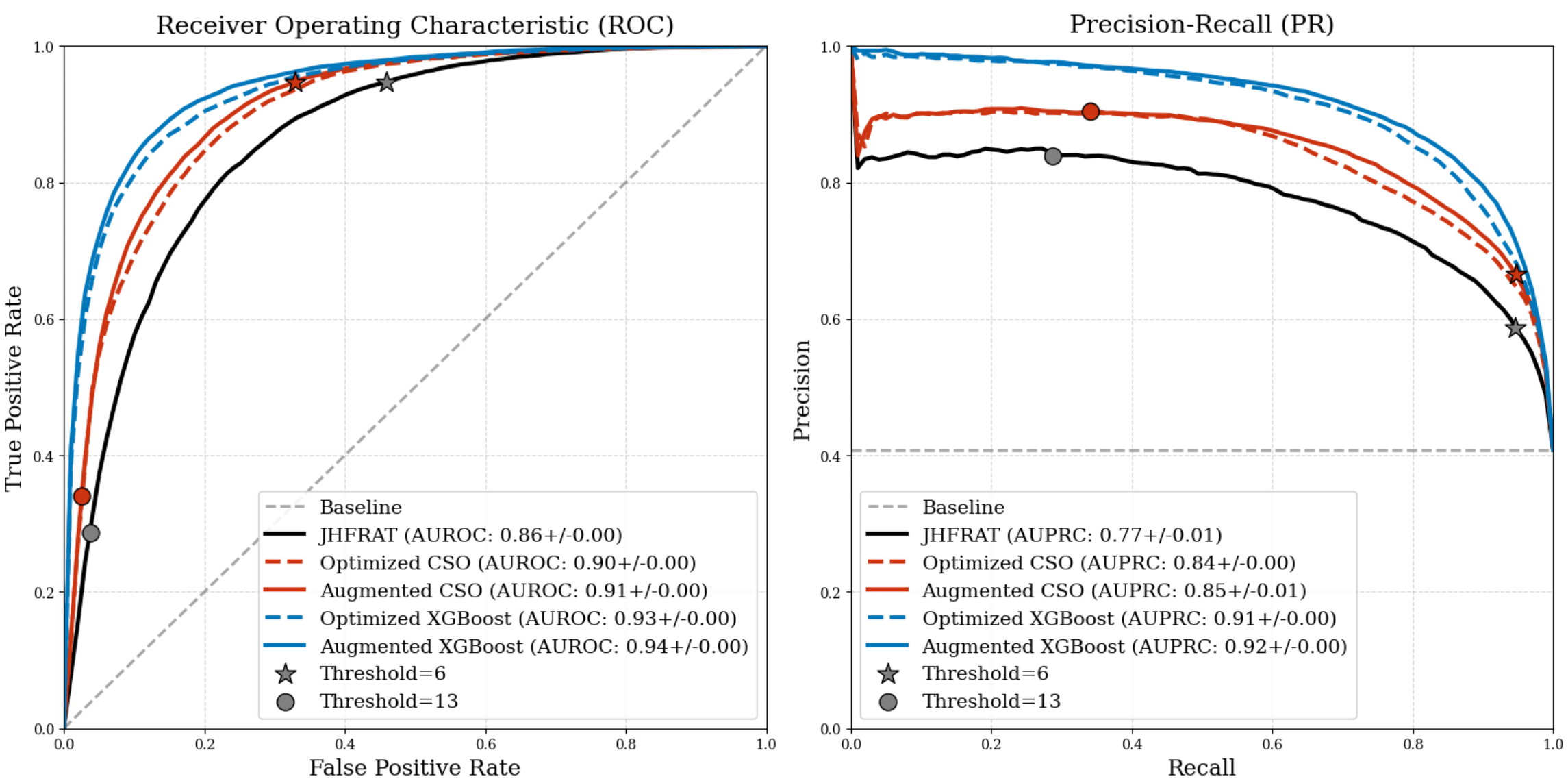

Figure 4. Receiver Operator Characteristic (ROC) and Precision-Recall (PR) curves for each model. Performance metrics from 5-fold cross-validation are reported as mean ± standard deviation. All models improve on JHFRAT, with augmented models (with EHR variables) marginally improving on the models without EHR variables.

Analysis of model performance demonstrated substantial improvements in fall risk prediction through optimization, with modest additional improvement from systematic integration of EHR variables. Figure 4 shows the receiver operator characteristic curves and precision-recall curves for each model. The augmented XGBoost (AUROC: 0.94, AUPRC: 0.92) and augmented CSO (AUROC: 0.91, AUPRC: 0.85) slightly outperform their optimized (without EHR variables) counterparts (AUROC: 0.93, AUPRC: 0.91 and AUROC 0.90, AUPRC: 0.84, respectively) and considerably outperform the current JHFRAT (AUROC: 0.86, AUPRC 0.77). With a high-risk threshold of 6, the augmented CSO models achieved a false positive rate of 0.33 compared to the original rate of 0.46 with the same true positive rate (0.95). With a high-risk threshold of 13, the augmented CSO model achieved modest improvements in both the false positive rate (CSO: 0.02, JHFRAT: 0.04) and true positive rate (CSO: 0.34, JHFRAT: 0.29). These gains reflect score recalibration and improved separation of low- vs high-risk labels under the existing JHFRAT category thresholds, compared to post-hoc probability calibration. While XGBoost achieves the best performance in ROC and PR metrics, the interpretability of the JHFRAT variables and model feature importance are less clear, as discussed in Section 3.3. and Table 1.

### 3.3. Feature Contribution and Sensitivity

| Feature Category | Variable | Occurrence Rate | Baseline: JHFRAT Coefficient | Optimized CSO Coefficient | Augmented CSO Coefficient | Optimized XGBoost SHAP | Augmented XGBoost SHAP |
|---|---|---|---|---|---|---|---|
| ***JHFRAT Variables*** | | | | | | | |
| ***Age*** | 60 - 69 years | 0.221 | 1 *(2.0%)* | 0.6 *(1.3%)* ↓ | 0.4 *(1.0%)* ↓ | -0.005 | -0.030 |
| | 70 - 79 years | 0.197 | 2 *(4.1%)* | 1.0 *(2.1%)* ↓ | 0.7 *(2.0%)* ↓ | -0.011 | |
| | Greater than or equal to 80 years | 0.132 | 3 *(6.1%)* | 1.4 *(2.8%)* ↓↓ | 1.0 *(2.8%)* ↓↓ | -0.001 | |
| ***Elimination, Bowel and Urine*** | Incontinence | 0.174 | 2 *(4.1%)* | 2.9 *(5.8%)* ↑ | 2.3 *(6.2%)* ↑ | -0.083 | -0.088 |
| | Urgency or frequency | 0.072 | 2 *(4.1%)* | 3.4 *(6.8%)* ↑ | 3.6 *(9.8%)* ↑↑ | -0.005 | -0.008 |
| ***Cognition*** | Altered awareness of immediate physical environment | 0.095 | 1 *(2.0%)* | 4.0 *(8.2%)* ↑↑ | 3.7 *(10.1%)* ↑↑ | 0.019 | 0.025 |
| | Impulsive | 0.029 | 2 *(4.1%)* | 4.0 *(8.2%)* ↑↑ | 3.7 *(10.1%)* ↑↑ | 0.055 | 0.046 |
| | Lack of understanding of one's physical and cognitive limitations | 0.047 | 4 *(8.2%)* | 4.0 *(8.2%)* | 3.7 *(10.1%)* ↑ | 0.034 | 0.026 |
| ***Patient Care Equipment*** | One present | 0.403 | 1 *(2.0%)* | 1.8 *(3.6%)* ↑ | 1.2 *(3.2%)* ↑ | -0.024 | -0.011 |
| | Two present | 0.213 | 2 *(4.1%)* | 2.7 *(5.5%)* ↑ | 2.1 *(5.8%)* ↑ | -0.021 | -0.025 |
| | Three or more present | 0.129 | 3 *(6.1%)* | 3.1 *(6.4%)* | 3.1 *(8.5%)* ↑ | 0.009 | 0.000 |
| ***Fall History*** | One fall within 6 months before admission | 0.119 | 5 *(10.2%)* | 0.0 *(0.0%)* ↓↓ | 0 *(0.0%)* ↓↓ | 0.002 | 0.008 |
| ***Medications*** | On 1 high fall risk drug | 0.337 | 3 *(6.1%)* | 2.0 *(4.2%)* ↓ | 0.8 *(2.2%)* ↓↓ | -0.021 | -0.016 |
| | On 2 or more high fall risk drugs | 0.407 | 5 *(10.2%)* | 2.4 *(4.9%)* ↓↓ | 1.2 *(3.4%)* ↓↓ | -0.010 | -0.012 |
| | Sedated procedure within past 24 hours | 0.032 | 7 *(14.3%)* | 2.4 *(4.9%)* ↓↓ | 1.3 *(3.7%)* ↓↓ | 0.004 | -0.001 |
| ***Mobility*** | Requires assistance | 0.511 | 2 *(4.1%)* | 6.5 *(13.3%)*↑↑ | 4.3 *(11.7%)* ↑↑ | 0.294 | -0.155 |
| | Unsteady gait | 0.080 | 2 *(4.1%)* | 3.8 *(7.7%)* ↑ | 2.0 *(5.4%)* ↑ | -0.010 | -0.012 |
| | Visual or auditory impairment affecting mobility | 0.015 | 2 *(4.1%)* | 3.0 *(6.1%)* ↑ | 1.4 *(3.8%)* ↓ | 0.004 | 0.004 |
| *JHFRAT Variables Coefficient Sum Subtotal* | | | 49 *(100%)* | 49 *(100%)* | 36.5 *(100%)* | N/A | N/A |
| ***Additional EHR Variables*** | | | | | | | |
| ***AMPAC*** | <= 25 | 0.091 | | | 2.6 | | -0.160 |
| | 25-35 | 0.094 | | | 4.1 | | |
| | 35-45 | 0.272 | | | 3.1 | | |
| | >45 | 0.416 | | | 0 | | |
| ***JHHLM*** | 1-3 | 0.180 | | | 0 | | -0.014 |
| | 4-5 | 0.081 | | | 0.3 | | |
| | 6-8 | 0.648 | | | 0.4 | | |
| ***Documented Comorbidities*** | <5 | 0.756 | | | 0 | | -0.001 |
| | 5-10 | 0.240 | | | 0.6 | | |
| | >10 | 0.004 | | | 0.4 | | |
| ***Sex*** | Female | 0.502 | | | 0 | | -0.001 |
| | Male | 0.498 | | | 0 | | 0.000 |
| ***Race*** | Black | 0.334 | | | 0.1 | | -0.001 |
| | White | 0.551 | | | 0 | | 0.000 |
| | Other | 0.115 | | | 0.2 | | 0.000 |
| ***Service Category*** | Medicine | 0.588 | | | 1.3 | | -0.007 |
| | Surgery | 0.179 | | | 0 | | 0.005 |
| | Oncology/Hematology | 0.071 | | | 0.4 | | 0.000 |
| | Neurosurgery | 0.053 | | | 2.9 | | -0.009 |
| | Orthopedics | 0.036 | | | 1.6 | | -0.007 |
| | Neurology | 0.031 | | | 2.4 | | 0.000 |
| | Other | 0.043 | | | 0 | | 0.003 |
| *Additional EHR Variables Coefficient Sum Subtotal* | | | 0 | 0 | 20.4 | N/A | N/A |
| *All Variables Coefficient Sum Total* | | | 49 | 49 | 56.81 | N/A | N/A |

Table 1: Comparison of variable contribution across models. The values for the items of the JHFRAT assessment are displayed as: coefficient (percentage of sum of coefficients for JHFRAT items only) to fairly compare feature importance. The occurrence rate, augmented CSO coefficients, and augmented XGBoost SHAP values are all reported as the average across cross-validation folds. To fairly compare the optimized and augmented feature importances, percentages included for the JHFRAT variables indicate the specific variable coefficient's portion of the total sum of JHFRAT variable coefficients, and arrows indicate the change in this percentage compared to the JHFRAT baseline.

Table 1 highlights the variable coefficients and their relative contribution between the baseline JHFRAT and each of the machine learning models. The relative feature importances for the JHFRAT variables do not vary significantly between the optimized and augmented models. Original JHFRAT components, such as cognition and mobility, retained substantial coefficients in the CSO models, reflecting their clinical relevance. The marginal improvement gained from the addition of EHR variables also suggests that the existing JHFRAT items capture most of the relevant risk information. However, some variables, such as 'One fall within 6 months before admission', do not appear as significant in CSO models, some that are linearly scored in JHFRAT, e.g., Mobility, appear more exponential in SCO, and some exponentially scored variables, e.g., Cognition, present more uniform score distributions. Notably, the percentage contributions of Age, Fall History, and Medication are reduced in CSO compared with JHFRAT, while Cognition, Patient Care Equipment, and Mobility have increased. Within the additional EHR variables, only AMPAC and the Service Categories in Neurosurgery, Orthopedics, and Neurology show notable contributions, and other variables, including Gender and Race, do not stand out.

Tree SHAP analysis provided insights into the relative importance of individual features within the XGBoost framework. Among these, the AM-PAC mobility score and the JHFRAT Requires Assistance mobility item emerged as the most influential features, maintaining consistency with their importance in the Augmented CSO model.

Table 2 shows how the cohort composition changes as we vary the minimum number of interventions needed per 3-day window for a patient to qualify as high-risk, and Figure 5 shows the feature importance variation between folds of the 5-fold cross-validation across these different cohort variations The constrained score optimization feature coefficients remain stable, while the XGBoost feature importances vary more substantially. The feature with the largest overall variation for CSO, altered awareness in the cognition category, varies by 2.88 percentage points, between a minimum of 5.27% and maximum of 8.15% of the total coefficient sum. The two features with the largest importance variation for the XGBoost model are the two with the highest importance overall: 'Requires Assistance' in the mobility category (11.00 – 35.15% of SHAP sum), and the AM-PAC score (4.58 – 22.71% of SHAP sum). These features become more and less important, respectively, as the portion of high-risk patients in the model cohort increases, as seen in Figure 6.

| **High-Risk Intervention Threshold** | **Number of Encounters per Risk Label** | | |
|---|---|---|---|
| | Low | High | Indeterminate |
| **4** | 19,762 | 20,527 | 6,747 |
| **5** | 20,073 | 16,813 | 10,150 |
| **6** | 20,208 | 13,941 | 12,887 |
| **7** | 20,390 | 11,441 | 15,205 |
| **8** | 20,454 | 9,507 | 17,075 |

Table 2: Risk labels across sensitivity analysis cohorts. While the low-risk population remains stable, high-risk population grows as the risk tolerance (# of interventions) is reduced.

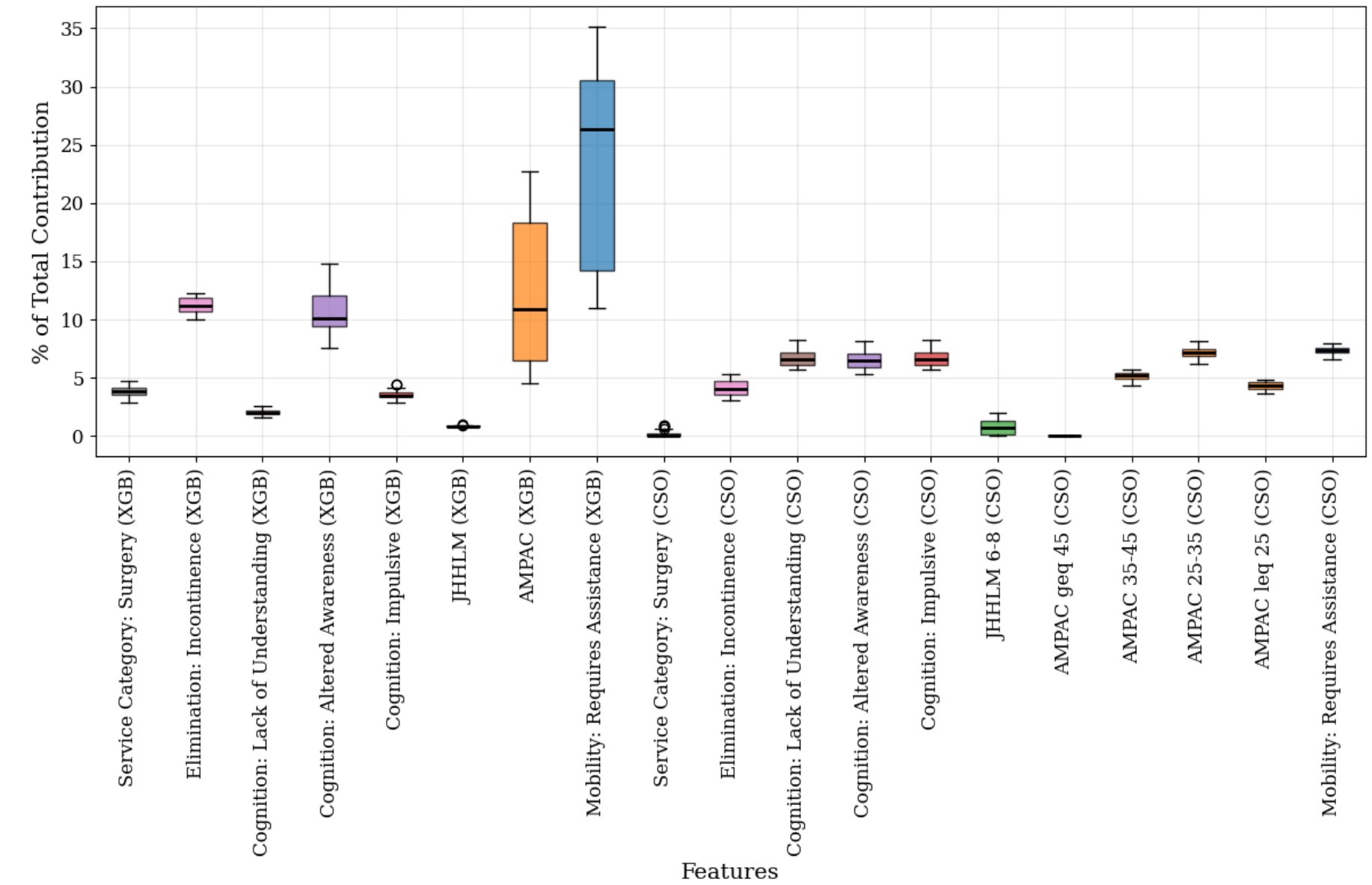

Figure 5 (Above): Feature Importance for five most-variable features (according to highest standard deviation) for each of the constrained logistic regression and XGBoost models. Percent total contribution to model importance is measured as feature coefficient divided by coefficient sum for the CSO model, and by mean absolute SHAP value divided by the total mean absolute SHAP values for all features. Feature importance is spread more evenly in CSO compared to XGBoost.

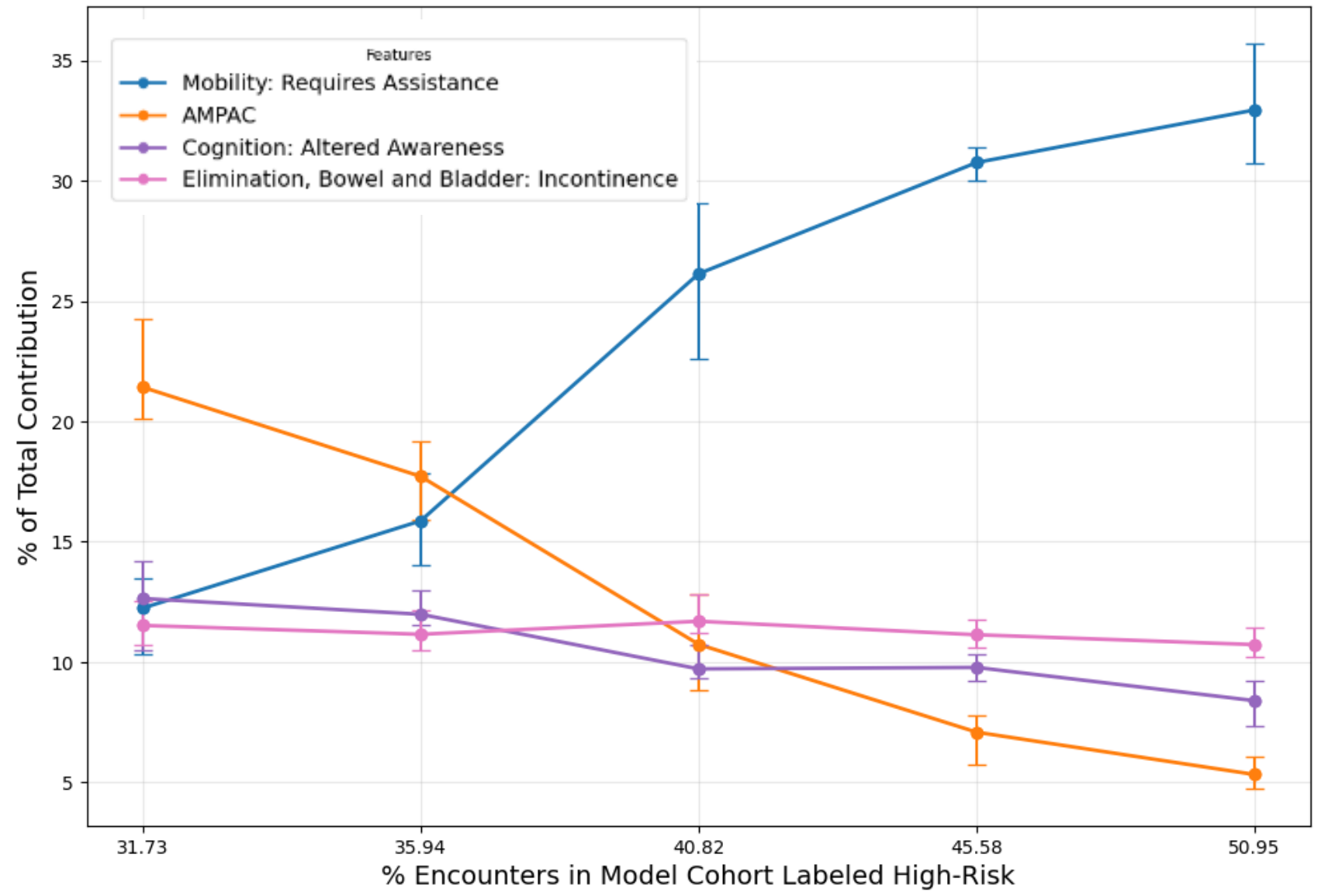

Figure 6 (Below): XGBoost feature importance over model cohort variations (per Table 3). Line plots show average, and error bars indicate minimum and maximum values across 5-fold cross-validation. JHFRAT Requires Mobility Assistance and AMPAC score show large variability as threshold for high-risk label is decreased.

### 3.4. Differential Risk Scoring Patterns

Analysis of risk score differentials between the augmented and baseline JHFRAT frameworks demonstrated structured modification of risk assessment patterns. Figure 7 shows how differential distributions exhibited approximately normal characteristics with slight positive asymmetry, suggesting stable augmentation of baseline risk scores without introducing substantial variability. The score differentials are more pronounced in the high-risk encounters, resulting in a higher portion of these patients moving from moderate risk to high risk (25.1 and 27.2 percentage points for optimized and augmented CSO, respectively) than from moderate to low risk (4.7 and 4.2 percentage points for optimized and augmented CSO, respectively). Both the optimized and augmented CSO models result in a reduction in the number of encounters in the study cohort considered moderate risk

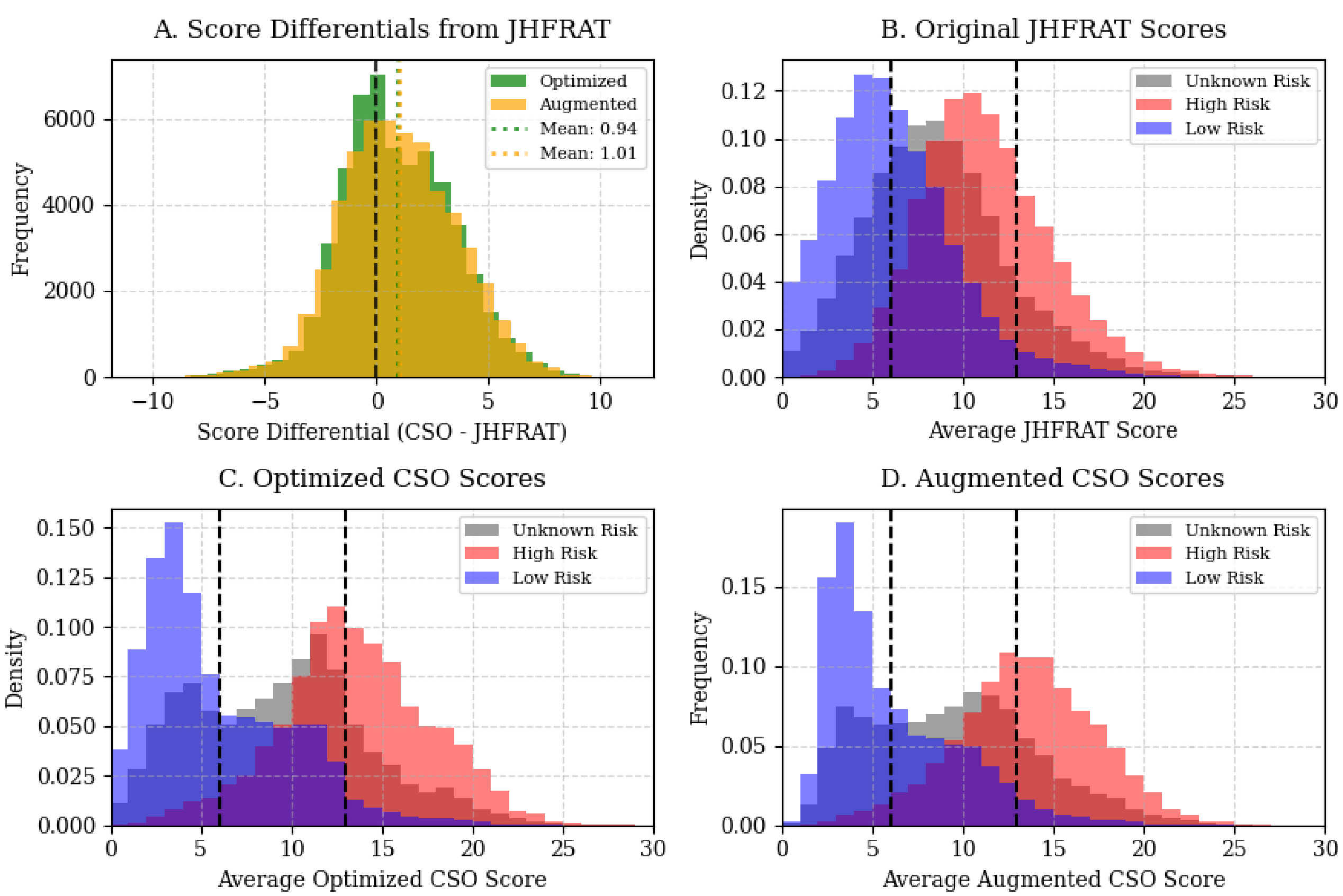

Figure 7: A) Score differentials demonstrate good alignment between each CSO model and JHFRAT, B-D) JHFRAT score distributions per class for the original JHFRAT, the optimized CSO, and the augmented CSO, respectively, for $\lambda = 0.5$. Augmented CSO (Fig D) shows the most separation between low- (blue) vs high-risk (red) patients.

## 4. Discussion

### 4.1. Clinical Impact and Patient Safety Outcomes

Data-driven optimization of the JHFRAT meaningfully improves fall risk stratification with immediate implications for patient safety and workflow efficiency. Table 3 displays the rates of classifying each type of patient (low-risk, high-risk, and unknown label) for the original JHFRAT and each CSO model. Our augmented CSO model correctly reclassified 3,788 high-risk patients (27% of the high-risk cohort) from moderate or low-risk categories, enabling proactive fall prevention interventions. Simultaneously, the model identified 860 low-risk patients (4.2% of the low-risk cohort) who were previously over-classified, potentially reducing unnecessary interventions and mobility restrictions that can lead to deconditioning and hospital-acquired complications[5]. A modest tradeoff exists: 264 low-risk patients (1.3%) were reclassified as high-risk.

In practical terms, for a 500-bed hospital with typical admission patterns similar to our study cohort, our model's enhanced risk stratification could protect an additional 35 high-risk patients per week through targeted interventions such as increased hourly rounding, toileting assistance, and mobility support, while only 2-3 additional low-risk patients per week would be unnecessarily classified as high-risk. Concurrently, approximately 8 low-risk patients per week could avoid unnecessary restrictions, promoting early mobilization and reducing hospital-acquired complications. This precision enables nurses to focus intensive interventions on a smaller, better-defined high-risk group rather than applying moderate-intensity interventions broadly. With average score differentials of 0.94 and 1.01 for the optimized and augmented CSO models, respectively, minimal total increase in targeted interventions can be expected.

| Binary Risk Label | Model | Patients in Predicted Risk Category (percent for risk label) | | |
|---|---|---|---|---|
| | | Low | Moderate | High |
| **Non-Fall Encounters** | | | | |
| Low Risk | Original | **11,271 *(56%)*** | 8,077 *(40%)* | 752 *(4%)* |
| | Optimized CSO | **12,218 *(61%)*** | 6,928 *(35%)* | 954 *(5%)* |
| | Augmented CSO | **12,128 *(60%)*** | 6,955 *(35%)* | 1,017 *(5%)* |
| High Risk | Original | 790 *(6%)* | 9,039 *(66%)* | **3,891 *(28%)*** |
| | Optimized CSO | 537 *(4%)* | 5,845 *(43%)* | **7,338 *(54%)*** |
| | Augmented CSO | 463 *(3%)* | 5,633 *(41%)* | **7,624 *(56%)*** |
| Unknown Risk | Original | 5,677 *(29%)* | 11,740 *(59%)* | 2,472 (12%) |
| | Optimized CSO | 5,729 *(28%)* | 10,023 *(50%)* | 4,137 *(21%)* |
| | Augmented CSO | 5,392 *(27%)* | 10,216 *(51%)* | 4,281 *(22%)* |
| **Fall Encounters** | | | | |
| Low Risk | Original | ***22 (20%)*** | 71 *(66%)* | 15 *(14%)* |
| | Optimized CSO | **29 *(27%)*** | 66 *(61%)* | 13 *(12%)* |
| | Augmented CSO | **25 *(23%)*** | 69 *(64%)* | 14 *(13%)* |
| High Risk | Original | 7 *(3%)* | 106 *(48%)* | **108 *(49%)*** |
| | Optimized CSO | 2 *(1%)* | 63 *(29%)* | **156 *(71%)*** |
| | Augmented CSO | 3 *(1%)* | 55 *(25%)* | **163 *(74%)*** |
| Unknown Risk | Original | 9 *(5%)* | 100 *(59%)* | 60 *(36%)* |
| | Optimized CSO | 10 *(6%)* | 84 *(50%)* | 75 *(44%)* |
| | Augmented CSO | 10 *(6%)* | 87 *(52%)* | 72 *(43%)* |

Table 3: Number of patients per risk label and predicted risk category across models. Optimized and augmented CSO models increase the number of correctly classified patients (indicated in bold) and decrease the number of patients in the study cohort classified as moderate risk. Bolded numbers constitute correctly classified encounters.

### 4.2. Interpretation of Key Findings

The integration of certain EHR variables, especially AM-PAC mobility scores and medical service category, demonstrated additional predictive value. As evidenced by SHAP analysis, these EHR-derived features capture complex and nonlinear interactions that are particularly well-suited to the high-dimensional XGBoost framework. Meanwhile, the fixed dual thresholds of the CSO model allow for contribution of these additional features without skewing the overall JHFRAT score scale. The demographic variables of sex and race were minimally significant across models, suggesting that fall risk does not depend on these factors. Notably, the preservation of established JHFRAT components within the augmented models reinforces their continued relevance in fall risk assessment. This finding suggests that data-driven enhancements can complement rather than replace traditional frameworks, providing a pathway for systematic improvements across diverse healthcare settings.

A key strength of this study is its dual focus on performance and interpretability. Additionally, the use of SHAP analysis provides transparency into the contributions of individual features, enhancing trust in model outputs. The study's large cohort size, spanning three hospitals and over 34,000 admissions, adds robustness to its findings. From a clinical perspective, the ability to more accurately identify high-risk patients has immediate implications for resource allocation and patient safety. The increased sensitivity achieved by the data-driven models alleviates documentation burden and alarm fatigue and enables earlier and more targeted interventions, potentially reducing fall incidence rates and associated healthcare costs.

### 4.3. White-Box vs Black-Box Modelling

Our comparison of interpretable (CSO) versus black-box (XGBoost) models addresses a critical implementation consideration in healthcare AI, where providers are more likely to trust understandable decision-support systems.[22] Many clinical decision-support systems originate as completely knowledge-based models in that they are developed from agreed-upon clinical expertise. On the other hand, data-based models may incorporate clinical knowledge, but ultimately rely on historical patient data for development and validation, resulting in a critical transparency gap.[23] Data-based "white-box" models are intuitive and implementable within existing clinical workflows but are often outperformed by "black-box" models that exploit complex variable relationships.[24] However, black-box models have advanced computational requirements and implementation barriers, necessitating careful consideration of the performance-interpretability tradeoff.

Table 4 highlights key differences in the CSO and XGBoost models. While the unconstrained XGBoost model achieves higher performance metrics, the CSO model demonstrates that substantial improvements over the current JHFRAT can be achieved while preserving interpretability and seamless integration into existing clinical workflows. A key advantage of the CSO approach is that it recalibrates the scores to better reflect the study's decision-informed risk labels, while maintaining the familiar JHFRAT scoring structure and risk category thresholds, ensuring that nurses can continue using established protocols without disruption. The optimization framework aligns the score distribution to current practice patterns while improving discriminative performance.

| | CSO | XGBoost |
|---|---|---|
| AUC-ROC | 0.91 | 0.94 |
| PR-ROC | 0.85 | 0.92 |
| Maximum feature importance range (percentage points) | 2.88 | 24.15 |
| Numerical Risk Score | Yes | |
| Ordinal classification adaptation | Yes | |
| Off-the-shelf | | Yes |

Table 4: Summary of augmented CSO and XGBoost performance metrics and requirements. XGBoost outperforms CSO in ROC metrics but is more sensitive to changes in risk label thresholds, as indicated by the higher feature importance range.

XGBoost offers "off-the-shelf" implementation but requires specialized expertise for deployment and ongoing maintenance. The CSO model requires clinical knowledge during initial development but results in a transparent, easily auditable scoring system that clinical staff can understand and trust, facilitating adoption.

### 4.4. Limitations and Future Directions

Several limitations warrant consideration for future implementation and research. The reliance on AM-PAC and JH-HLM mobility assessments may limit immediate generalizability to institutions without these standardized mobility measures. However, alternative mobility indicators routinely collected in EHRs (such as physical therapy assessments or ambulation orders) could potentially substitute for these variables with appropriate validation. Furthermore, use of sensor data, including from wearables, as a contributing factor in the risk assessment could help provide additional predictive power, as such data has been shown valuable for fall prediction in both outpatient and inpatient settings.[25-27]

Our retrospective design and intervention-based risk labeling approach, while methodologically sound, represents an indirect, decision-informed measure of fall risk that reflects clinically signaled risk through preventive care patterns rather than intrinsic biologic susceptibility. Future prospective studies could validate these findings in controlled settings and explore real-time risk prediction models that continuously update patient risk scores based on changing clinical status throughout hospitalization.

The study's focus on static risk assessment also suggests opportunities for dynamic risk modeling that incorporates temporal changes in patient condition, medication effects, and response to interventions. Such models could provide even more precise risk stratification and intervention timing recommendations.

Future validation studies should examine the practical feasibility of translating these optimized coefficients into clinical practice, including the technical challenges of EHR integration and the organizational factors that influence adoption of modified risk assessment tools. Additionally, prospective evaluation of clinical outcomes following implementation would be essential to confirm the predicted improvements in fall prevention and resource allocation observed in our retrospective analysis.

## References

1 Heikkilä A, Lehtonen L, Junttila K. Fall rates by specialties and risk factors for falls in acute hospital: A retrospective study. *Journal of Clinical Nursing*. 2023;**32(15-16):**4868-4877.

2 Wong CA, Recktenwald AJ, Jones ML, Waterman BM, Bollini ML, Dunagan WC. The cost of serious fall-related injuries at three Midwestern hospitals. *Jt Comm J Qual Patient Saf*. 2011;**37(2):**81-87.

3 Morello RT, Barker AL, Watts JJ, et al. The extra resource burden of in-hospital falls: a cost of falls study. *Med J Aust*. 2015;**203(9):**367

4 Kalisch BJ, Lee S, Dabney BW. Outcomes of inpatient mobilization: a literature review. *J Clin Nurs*. 2014;**23(11-12):**1486-1501.

5 Capo-Lugo CE, Young DL, Farley H, et al. Revealing the tension: The relationship between high fall risk categorization and low patient mobility. *J Am Geriatr Soc*. 2023;**71(5):**1536-1546.

6 Poe SS, Cvach M, Dawson PB, Straus H, Hill EE. The Johns Hopkins Fall Risk Assessment Tool: postimplementation evaluation. *J Nurs Care Qual*. 2007;**22(4):**293-298.

7 Poe SS, Cvach MM, Gartrell DG, Radzik BR, Joy TL. An Evidence-based Approach to Fall Risk Assessment, Prevention, and Management: Lessons Learned. *Journal of Nursing Care Quality*. 2005;**20(2):**107.

8 Lindberg DS, Prosperi M, Bjarnadottir RI, et al. Identification of important factors in an inpatient fall risk prediction model to improve the quality of care using EHR and electronic administrative data: A machine-learning approach. *Int J Med Inform*. 2020;**143:**104272.

9 Shim S, Yu JY, Jekal S, et al. Development and validation of interpretable machine learning models for inpatient fall events and electronic medical record integration. *Clin Exp Emerg Med*. 2022;**9(4):**345-353.

10 Kang CW, Yan ZK, Tian JL, Pu XB, Wu LX. Constructing a fall risk prediction model for hospitalized patients using machine learning. *BMC Public Health*. 2025;**25(1):**242.

11 Liu CH, Hu YH, Lin YH. A Machine Learning-Based Fall Risk Assessment Model for Inpatients. *Comput Inform Nurs*. 2021;**39(8):**450-459.

12 Wang HH, Huang CC, Talley PC, Kuo KM. Using Healthcare Resources Wisely: A Predictive Support System Regarding the Severity of Patient Falls. *J Healthc Eng*. 2022;**2022:**3100618.

13 Ladios-Martin M, Cabañero-Martínez MJ, Fernández-de-Maya J, et al. Development of a predictive inpatient falls risk model using machine learning. *J Nurs Manag*. 2022;**30(8):**3777-3786

14 Collins GS, Reitsma JB, Altman DG, Moons KGM. Transparent reporting of a multivariable prediction model for individual prognosis or diagnosis (TRIPOD): the TRIPOD statement. *BMJ*. 2015;**350:**g7594.

15 Jette DU, Stilphen M, Ranganathan VK, Passek SD, Frost FS, Jette AM. Validity of the AM-PAC “6-Clicks” inpatient daily activity and basic mobility short forms. *Phys Ther*. 2014;**94(3):**379-391.

16 Klein LM, Young D, Feng D, et al. Increasing patient mobility through an individualized goal-centered hospital mobility program: A quasi-experimental quality improvement project. *Nursing Outlook*. 2018;**66(3):**254-262.

17 Hu, D, Shi, X, Sun, L, et al. Comorbidity increased the risk of falls in Chinese older adults: A cross-sectional study. *Int J Clin Exp Med* 2017;**10(7):**10753-10763

18 Alshehri M, Alqahtani B, Alenazi A, Waitman L, Kluding P. Comorbidities and Medications Associated With Falls in Older Adults With Osteoarthritis: A Retrospective Study. *Archives of Physical Medicine and Rehabilitation*. 2019;**100(10):**e55

19 Hoyer E, Young D, Ke V, et al. Association of Longitudinal Mobility Levels in the Hospital and Injurious Inpatient Falls. *Am J Phys Med Rehabil*. 2024;**103(3):**251-255.

20 Kissane H, Knowles J, Tanzer JR, et al. Relationship Between Mobility and Falls in the Hospital Setting. *J Brown Hosp Med*. 2023;**2(3):**82146.

21 [preprint] Ganjkhanloo F, Springer E, Hoyer EH, Young DL, Ghobadi K. Joint Score-Threshold Optimization for Interpretable Risk Assessment Under Partial Supervision. *arXiv*. Preprint posted online October 24, 2025.21 Schwartz JM, George M, Rossetti SC, et al. Factors Influencing Clinician Trust in Predictive Clinical Decision Support Systems for In-Hospital Deterioration: Qualitative Descriptive Study. *JMIR Hum Factors*. 2022;**9(2):**e33960.

22 Mehta V, Komanduri A, Bhadouriya RS, et al. Evaluating transparency in AI/ML model characteristics for FDA-reviewed medical devices. *npj Digit Med*. 2025;**8(1):**673.

23 Schwartz JM, George M, Rossetti SC, et al. Factors Influencing Clinician Trust in Predictive Clinical Decision Support Systems for In-Hospital Deterioration: Qualitative Descriptive Study. *JMIR Hum Factors*. 2022;**9(2):**e33960.

24 Xu Q, Xie W, Liao B, et al. Interpretability of Clinical Decision Support Systems Based on Artificial Intelligence from Technological and Medical Perspective: A Systematic Review. *J Healthc Eng*. 2023;**2023:**9919269.

25 Bonanno M, Ielo A, Pasquale PD, et al. Use of Wearable Sensors to Assess Fall Risk in Neurological Disorders: Systematic Review. *JMIR mHealth and uHealth*. 2025;**13(1):**e67265.

26 Greene BR, McManus K, Redmond SJ, Caulfield B, Quinn CC. Digital assessment of falls risk, frailty, and mobility impairment using wearable sensors. *npj Digit Med*. 2019;**2(1):**125.

27 Sotirakis C, Brzezicki MA, Patel S, Conway N, FitzGerald JJ, Antoniades CA. Predicting future fallers in Parkinson’s disease using kinematic data over a period of 5 years. *npj Digit Med*. 2024;**7(1):**345.